\documentclass[letterpaper, 10 pt, conference]{ieeeconf}  

\IEEEoverridecommandlockouts                              
\usepackage[utf8]{inputenc}
\usepackage[T1]{fontenc}
\usepackage{amsmath,graphicx, adjustbox, inputenc}
\usepackage{array}
\usepackage{geometry}
\usepackage{booktabs}
\usepackage{longtable}
\usepackage{tikz}
\usetikzlibrary{shapes,arrows}
\usepackage{multirow}

\usepackage{subcaption}  
\usepackage{graphicx}    

\title{\LARGE \bf
Predicting Overtakes in Trucks Using CAN Data
}

\author{Talha Hanif Butt$^{1}$, Prayag Tiwari$^{1}$ and Fernando Alonso-Fernandez$^{1}$
\thanks{$^{1}$Talha Hanif Butt, Prayag Tiwari and Fernando Alonso-Fernandez is with  School of Information Technology, Halmstad University, Sweden
        {\tt\small talha-hanif.butt at hh.se, prayag.tiwari at
 hh.se, fernando.alonso-fernandez at hh.se}}%
}

\begin{document}

\maketitle
\thispagestyle{empty}
\pagestyle{empty}

\begin{abstract}

Safe overtakes in trucks are crucial to prevent accidents, reduce congestion, and ensure efficient traffic flow, making early prediction essential for timely and informed driving decisions. Accordingly, we investigate the detection of truck overtakes from CAN data. Three classifiers, Artificial Neural Networks (ANN), Random Forest, and Support Vector Machines (SVM), are employed for the task. Our analysis covers up to 10 seconds before the overtaking event, using an overlapping sliding window of 1 second to extract CAN features. We observe that the prediction scores of the overtake class tend to increase as we approach the overtake trigger, while the no-overtake class remain stable or oscillates depending on the classifier. Thus, the best accuracy is achieved when approaching the trigger, making early overtaking prediction challenging. The classifiers show good accuracy in classifying overtakes (Recall/TPR $\ge$ 93\%), but accuracy is suboptimal in classifying no-overtakes (TNR typically 80-90\% and below 60\% for one SVM variant). We further combine two classifiers (Random Forest and linear SVM) by averaging their output scores. The fusion is observed to improve no-overtake classification (TNR $\ge$ 92\%) at the expense of reducing overtake accuracy (TPR). However, the latter is kept above 91\% near the overtake trigger. Therefore, the fusion balances TPR and TNR, providing more consistent performance than individual classifiers.

\end{abstract}

\section{INTRODUCTION}

The development of Advanced Driver Assistance Systems (ADAS) has emerged as one of the most popular areas of research in artificial intelligence.
Through several sensors, ADAS is designed to alert the driver of potential hazards or control the vehicle to ultimately avoid collisions or accidents.
%
%
For those tasks, the vehicle must gather information about its surroundings to decide what to do and how to do it. 
Knowing the driver's intention is an integral part of the system, to determine if the ADAS should activate, providing opportune aids or alerts, or even overriding the driver's inputs \cite{c1}.

Among the most important driving manoeuvres is the overtaking manoeuvre in particular. 
Lane changes, acceleration and deceleration, and estimation of the speed and distance of the vehicle ahead or in the lane it is travelling in are all part of the process. 
Though there is a lot of work in the literature that aims at predicting driving manoeuvres, very few address overtaking \cite{c2,c3,c4}, and no real-world dataset is available due to the risk associated with overtaking \cite{c5}.
Most works address the estimation of lane change \cite{c1} or turning intention at intersections \cite{c6}.
In doing so, different data sources are typically used, including information from the driver (via cameras or biosensors capturing EEG, ECG, etc.), from the vehicle (CAN bus signals), or the traffic (GPS position or relative position or velocity of surrounding vehicles via cameras or lidar).

In this paper, we present ongoing work on overtake detection, in particular for trucks. 
Trucks carry heavier loads than cars, so a truck accident can be way more devastating. 
Accidents involving trucks can also lead to traffic congestion and delays due to their bigger size, and economic losses due to cargo being transported.
Ensuring driving security for trucks is thus crucial, especially when compared to lighter vehicles like cars.
We perform the task via CAN bus signals. We favour such signals because they are readily available onboard without the need for additional hardware like cameras or biosensors. This also avoids privacy concerns related to cameras looking inside or outside the cabin, or sensors capturing data from the driver.
We employ real CAN data from real operating trucks provided by Volvo Group participating in this research.
The contribution of this paper is that, to the best of our knowledge, we are the first to study overtake detection in trucks, particularly from real CAN bus data.
%
%
We also demonstrate that the fusion of classifiers can help to obtain a balanced performance in detecting the two classes (overtake, no overtake).

\begin{table}

\caption{Files employed per truck and class for training and testing. t1, t2, t3 denotes truck1, truck2 and truck3, respectively. class0=no overtake. class1=overtake.}

\label{tab0}

\resizebox{0.48\textwidth}{!}{\begin{tabular}{|l||l|l|l|l||l|l|l|l|}

\cline{2-9}

\multicolumn{1}{c||}{} &  \multicolumn{4}{|c||}{\textbf{class0}} &  \multicolumn{4}{|c|}{\textbf{class1}} \\ \cline{2-9}

\multicolumn{1}{c||}{} &  t1 & t2 & t3 & total &  t1 & t2 & t3 & total \\ \hline

\textbf{train} &  74 & 38 & 4 & 116 & 74 & 38 & 4 & 116 \\  \hline
 
\textbf{test} &  33 & 113 & 2 & 148 & 312 & 17 & 3 & 332 \\

\hline

\end{tabular}}
\end{table}

\begin{figure*}[h]
\caption{Boxplot of scores towards class0 (left column, no overtake class) and class1 (right, overtake) from -10 to +1 seconds around the trigger. From top to bottom row: ANN, RF, SVM linear and SVM rbf classifiers.}
\label{fig:boxplots}
\centering

\includegraphics[width=0.38\textwidth]{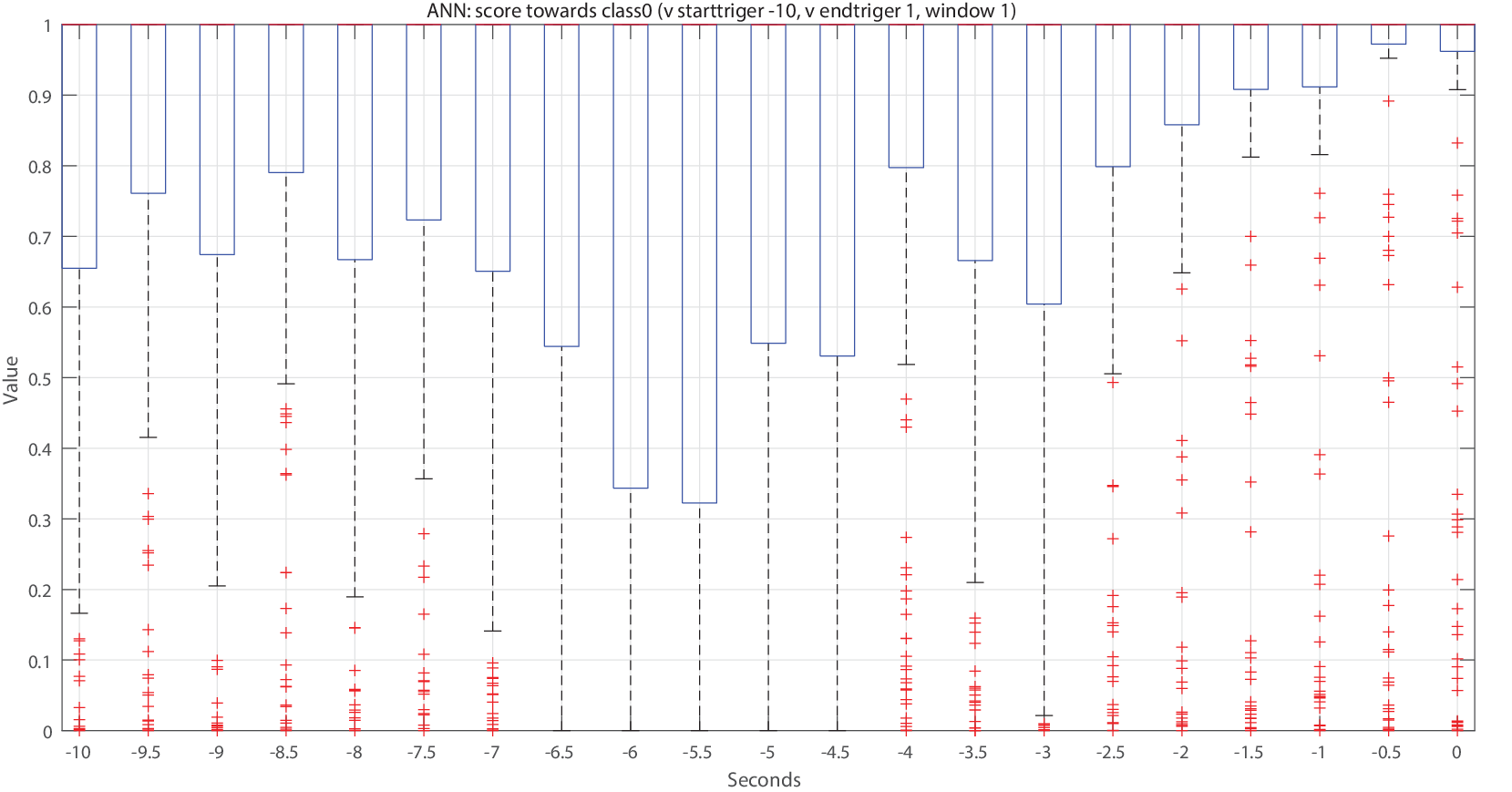}
\includegraphics[width=0.38\textwidth]{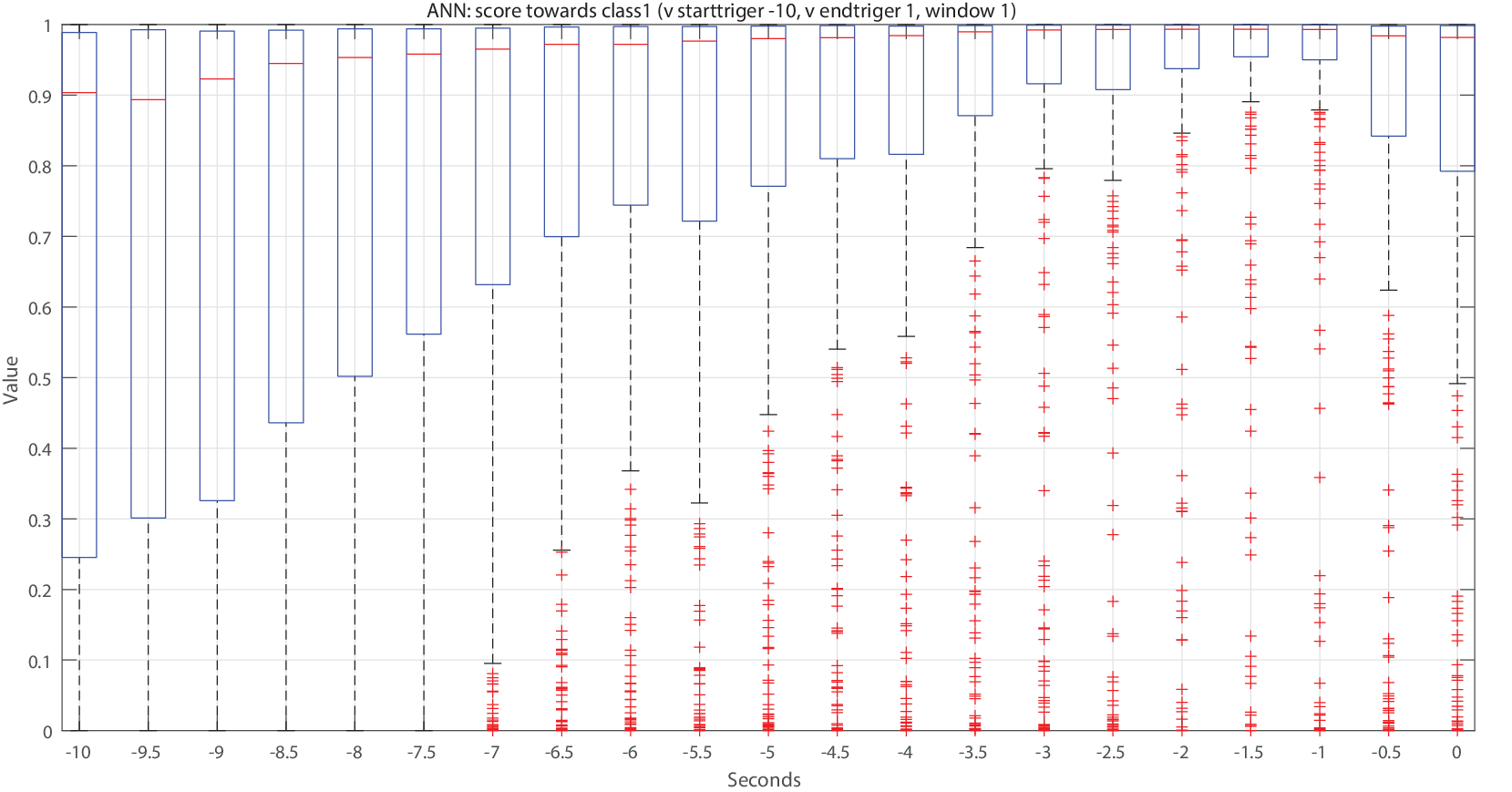}

\includegraphics[width=0.38\textwidth]{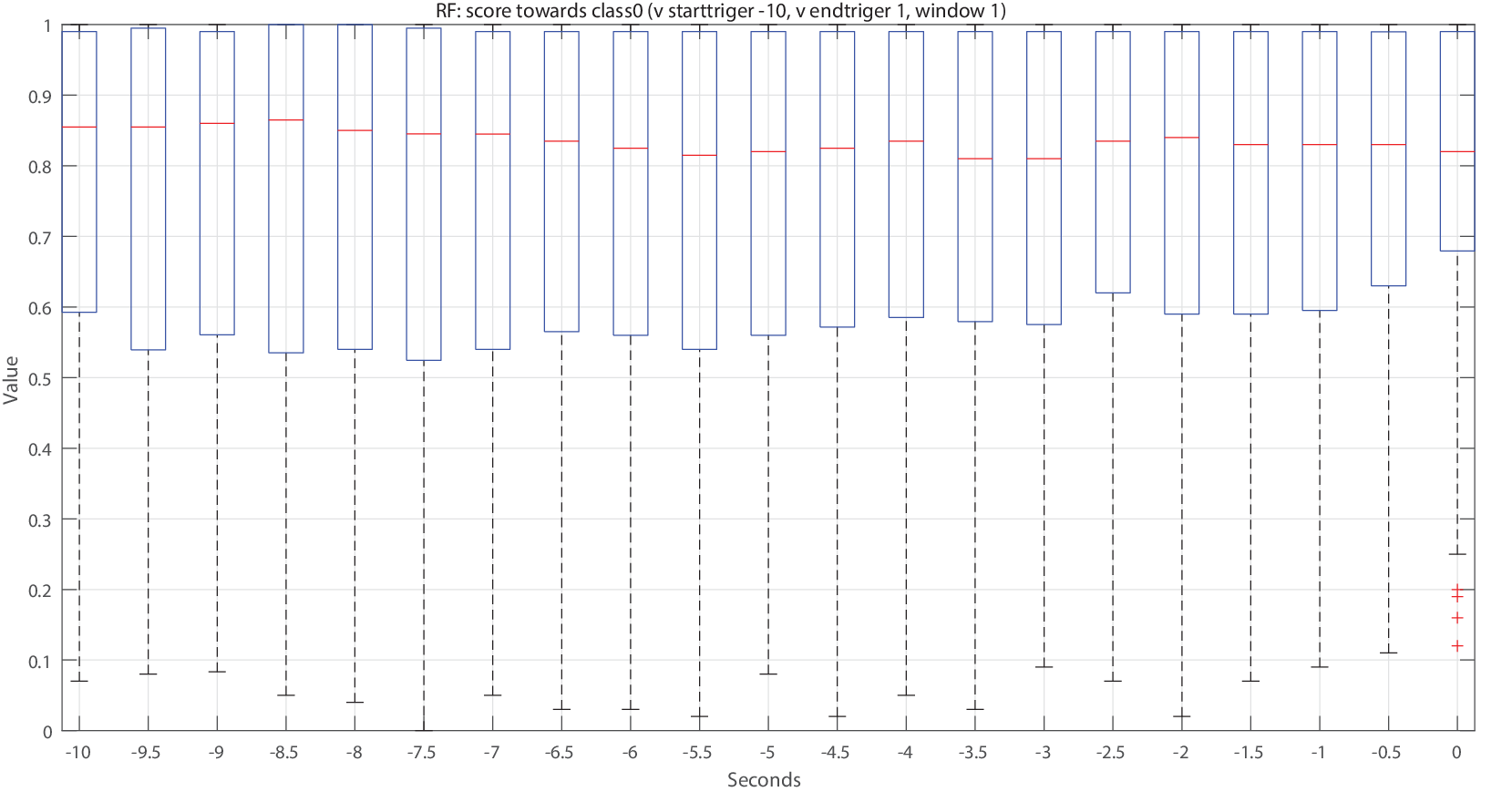}
\includegraphics[width=0.38\textwidth]{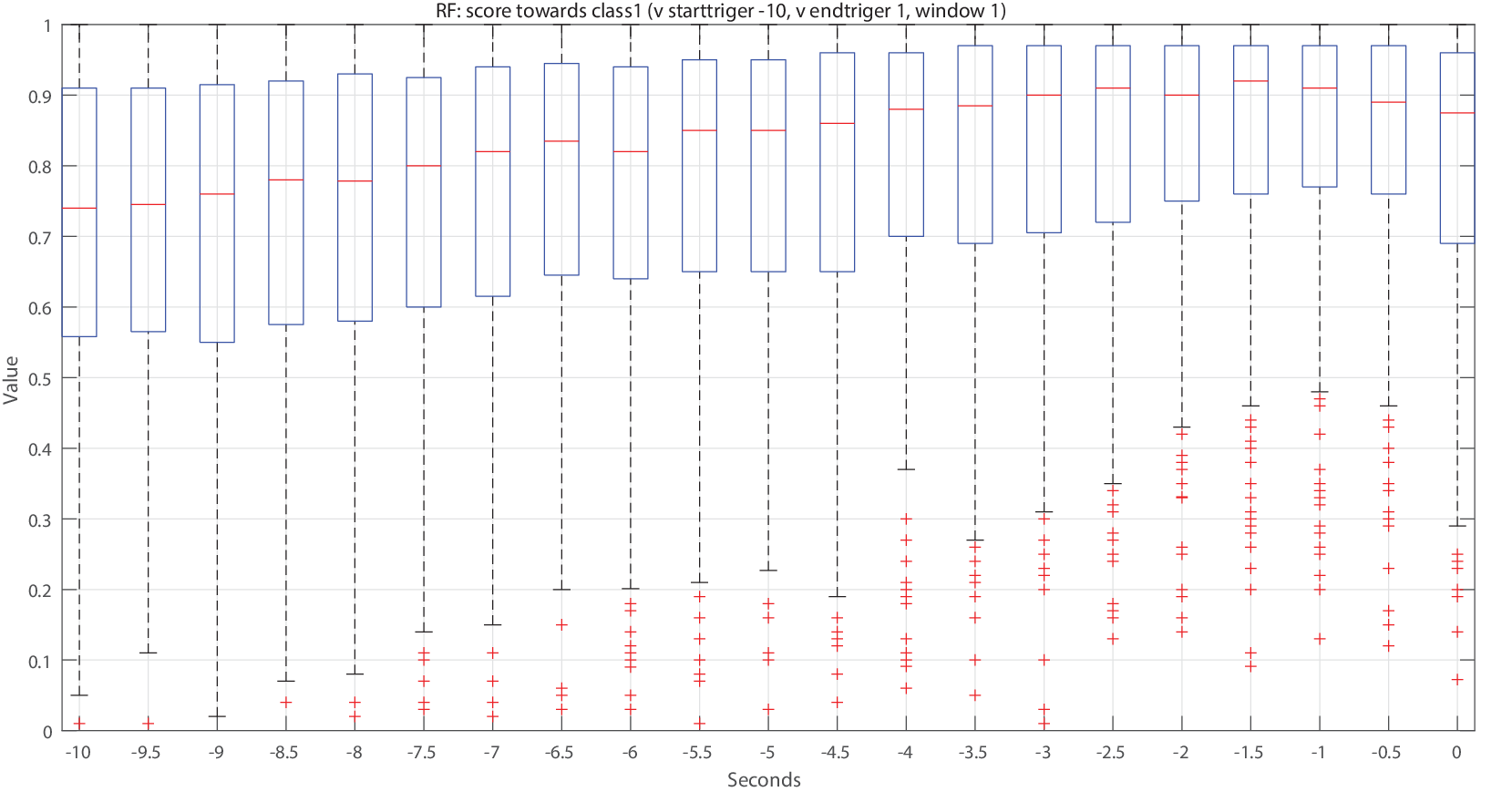}

\includegraphics[width=0.38\textwidth]{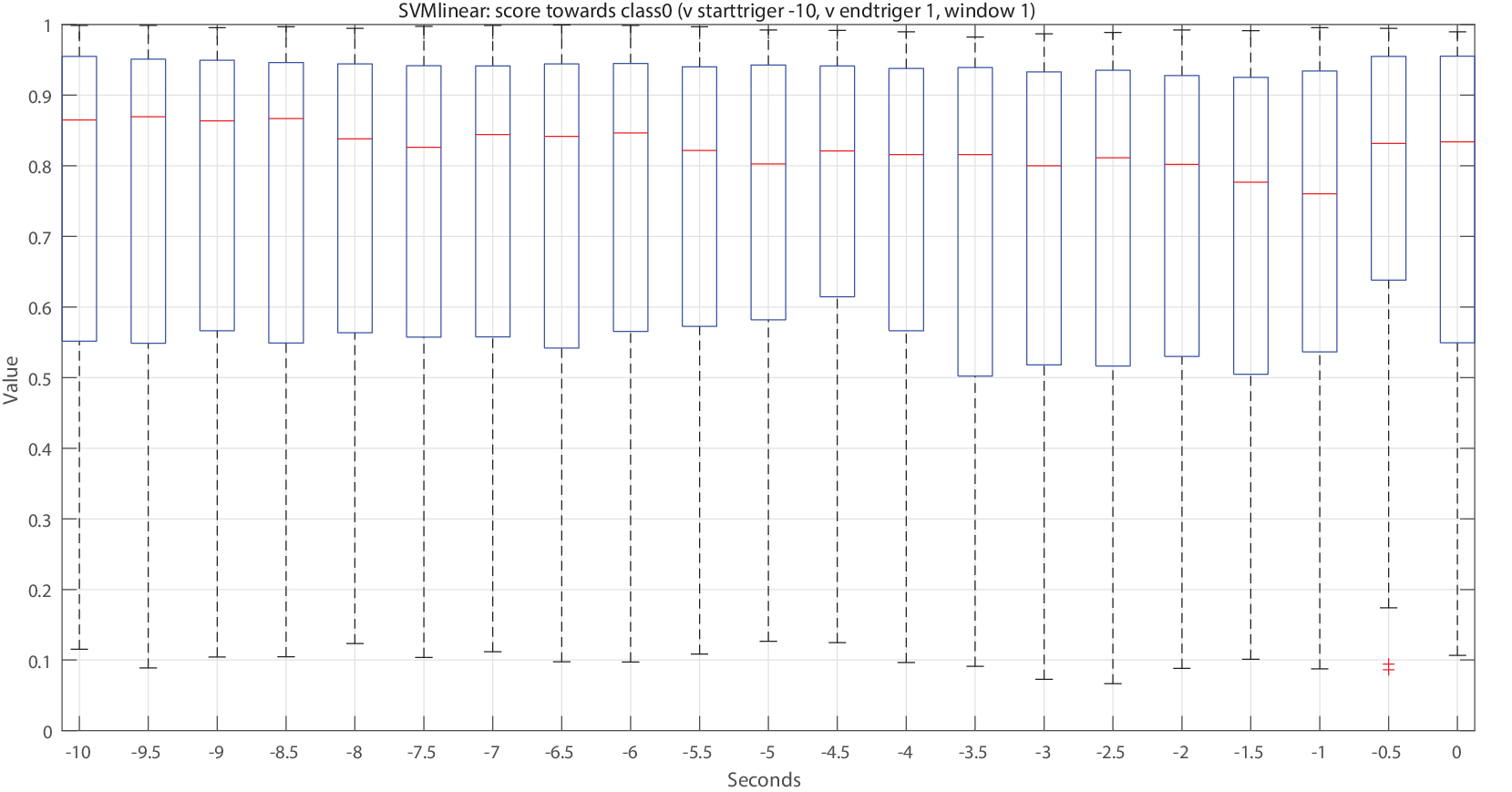}
\includegraphics[width=0.38\textwidth]{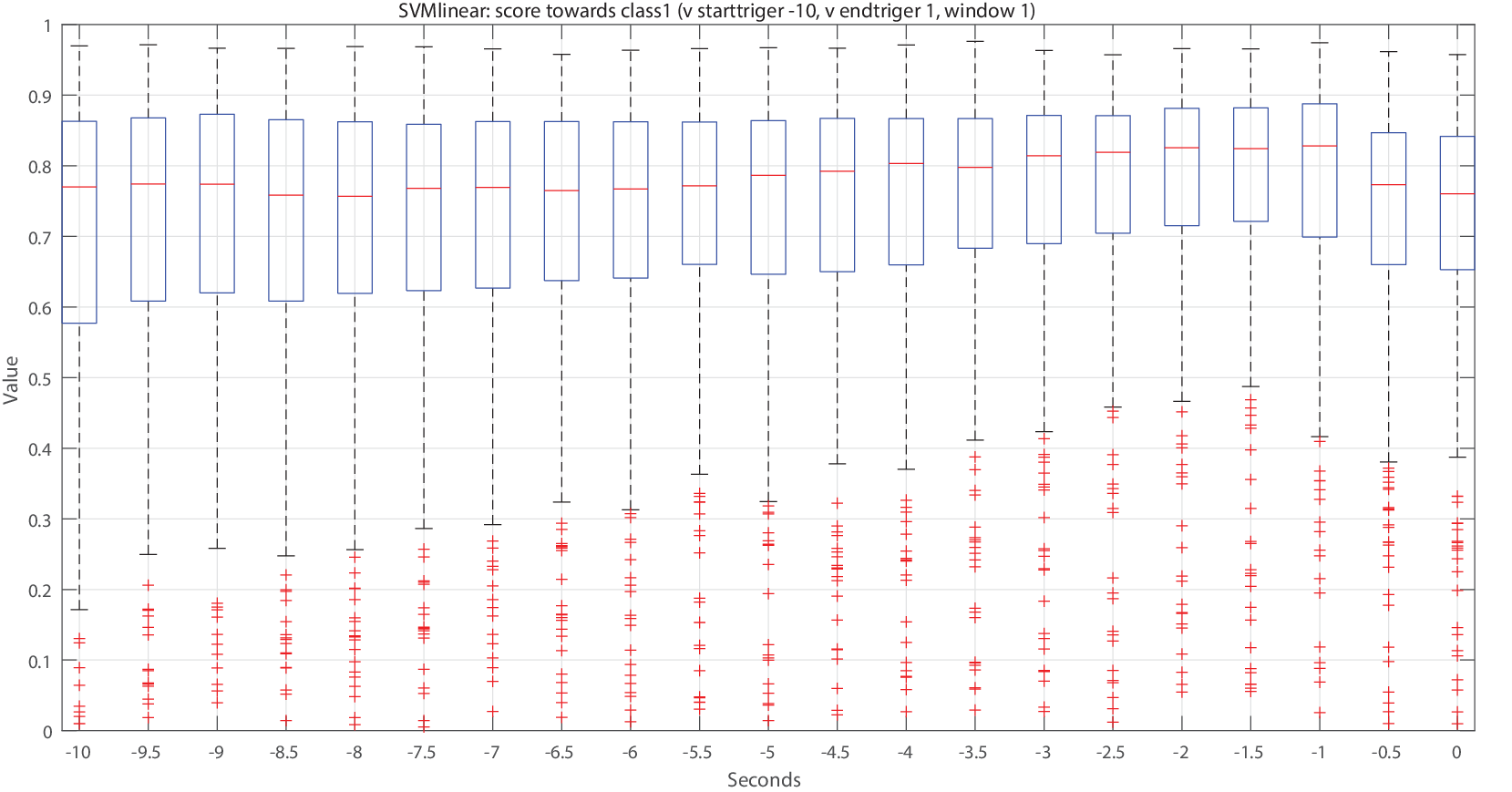}

\includegraphics[width=0.38\textwidth]{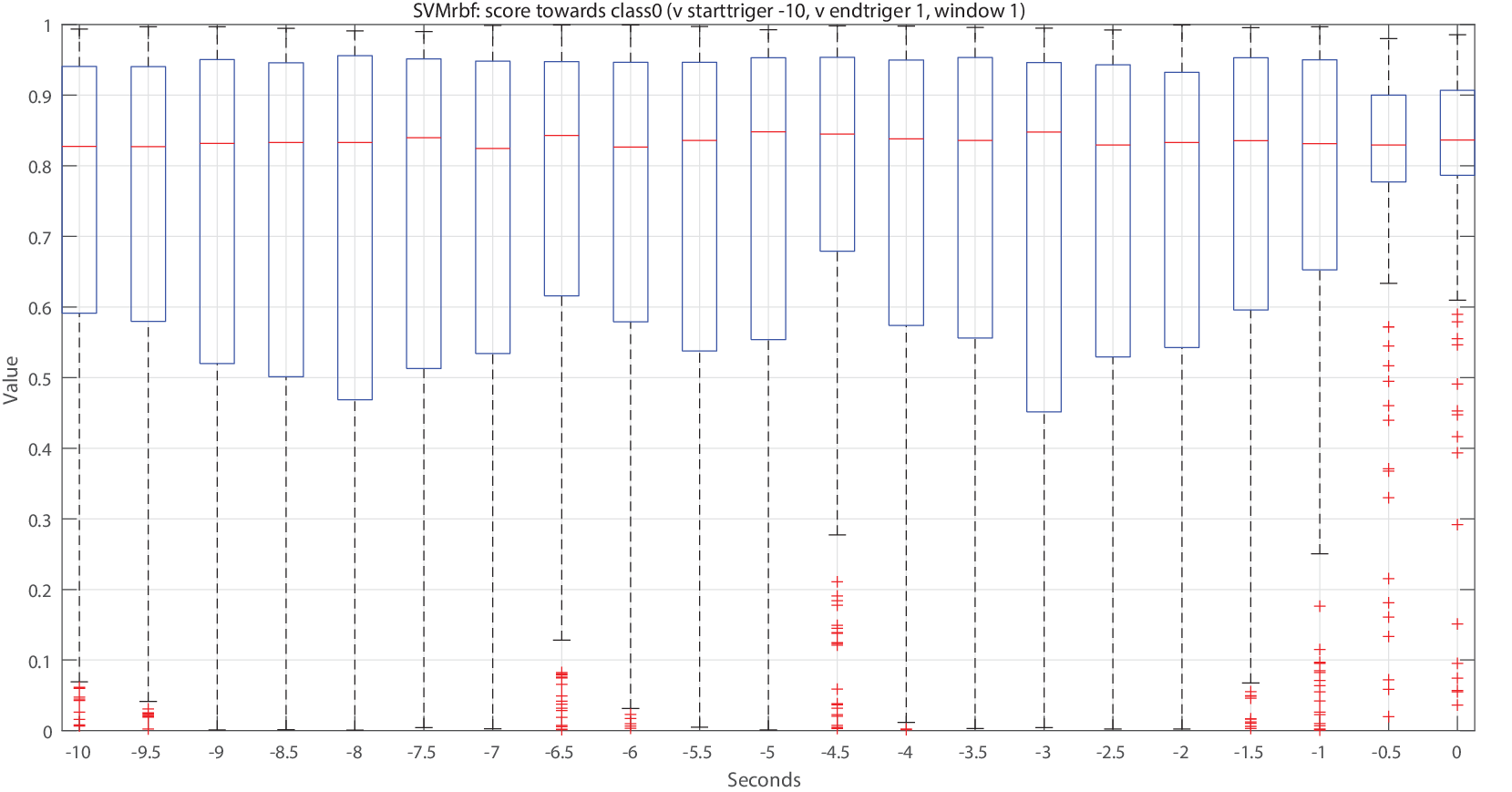}
\includegraphics[width=0.38\textwidth]{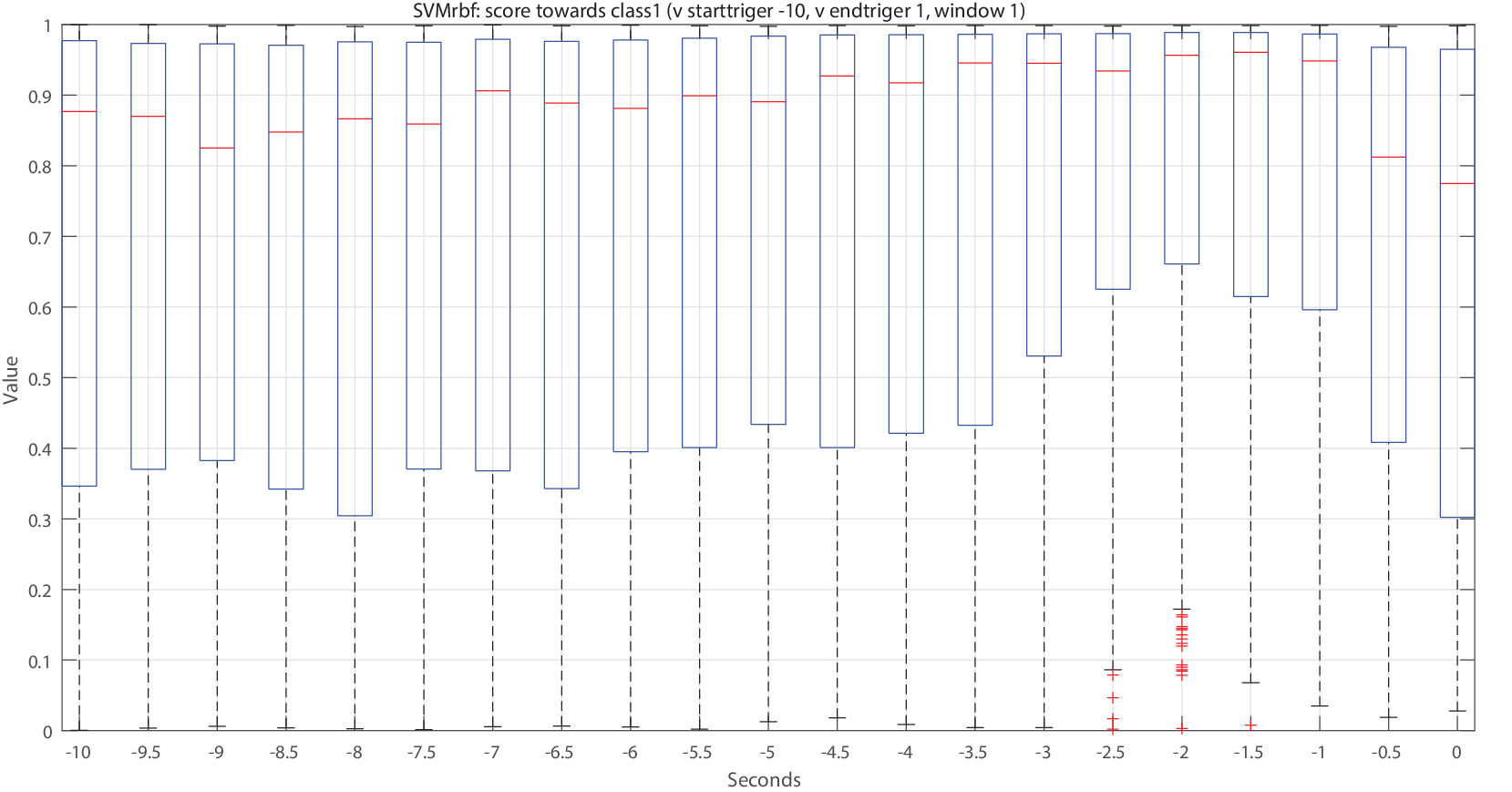}

\end{figure*}

\begin{figure*}[h]
\caption{Precision-Recall curves of the classifiers at different moments before the overtake maneuver starts. AUC (Area under the curve) values are given in Table~\ref{tab1}.}
\label{fig:PRcurves}
\centering

\includegraphics[width=0.24\textwidth]{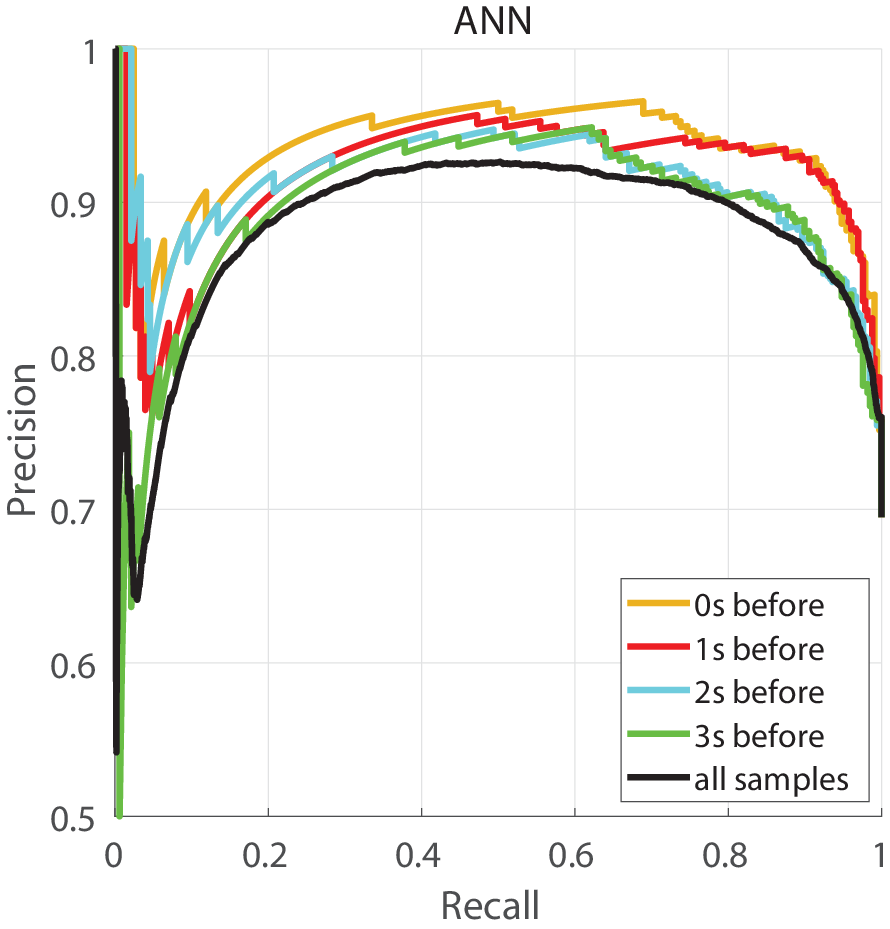}
\includegraphics[width=0.24\textwidth]{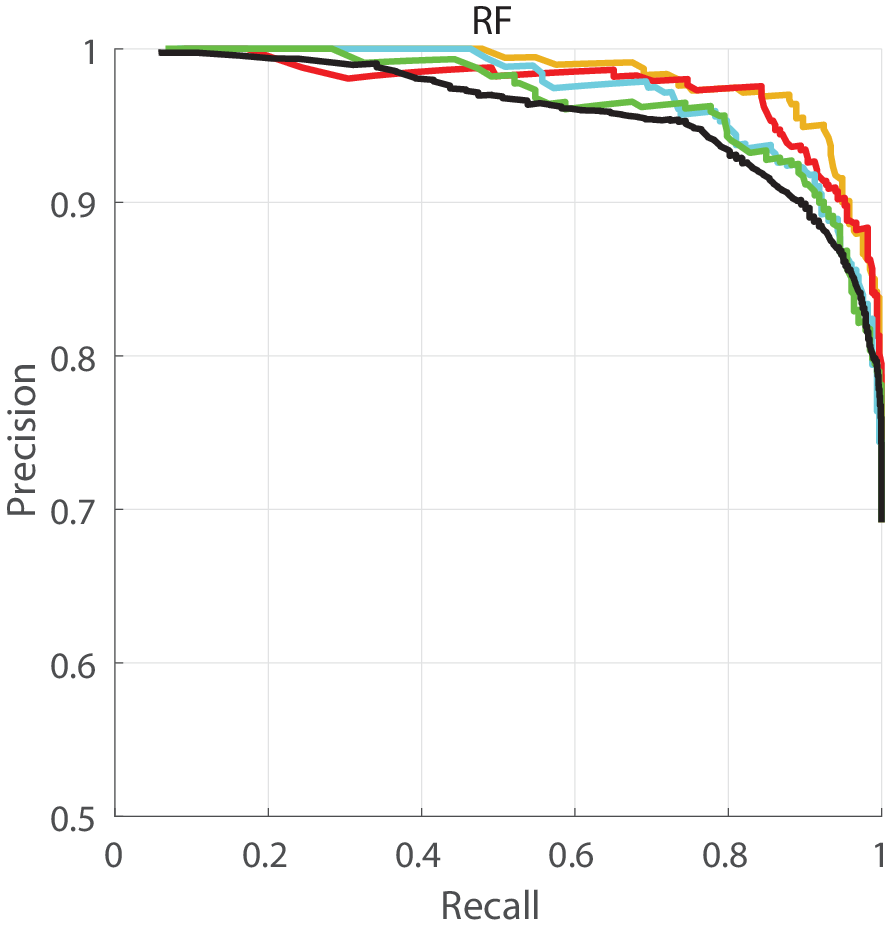}
\includegraphics[width=0.24\textwidth]{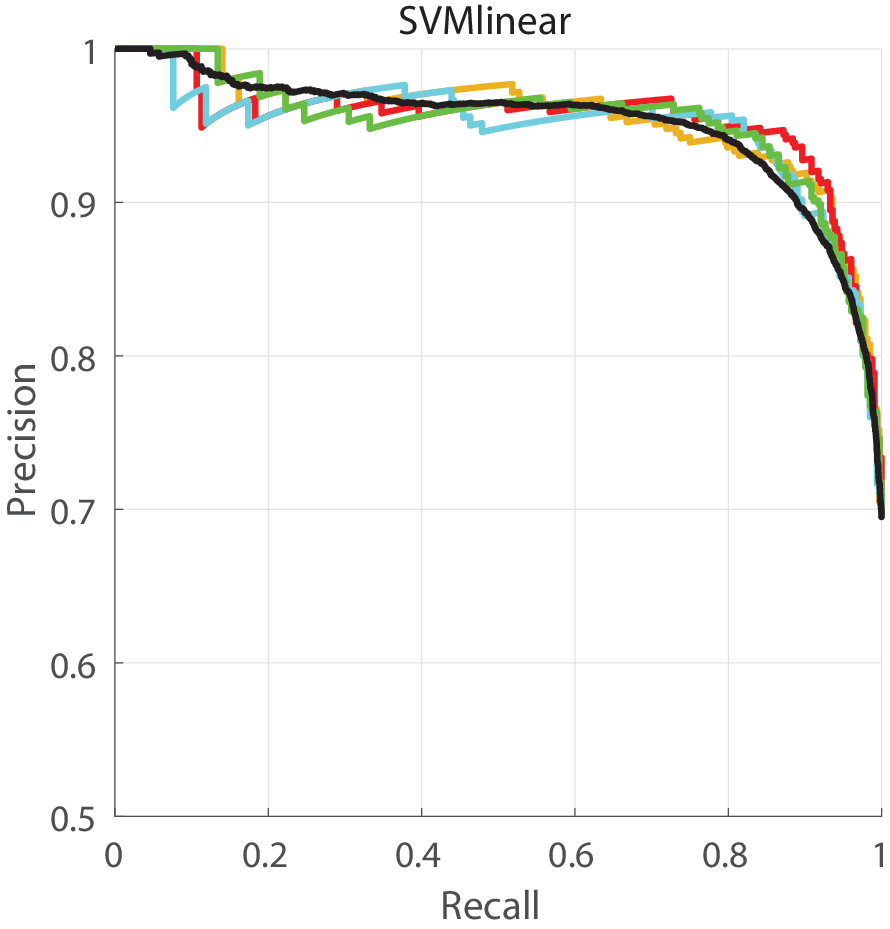}
\includegraphics[width=0.24\textwidth]{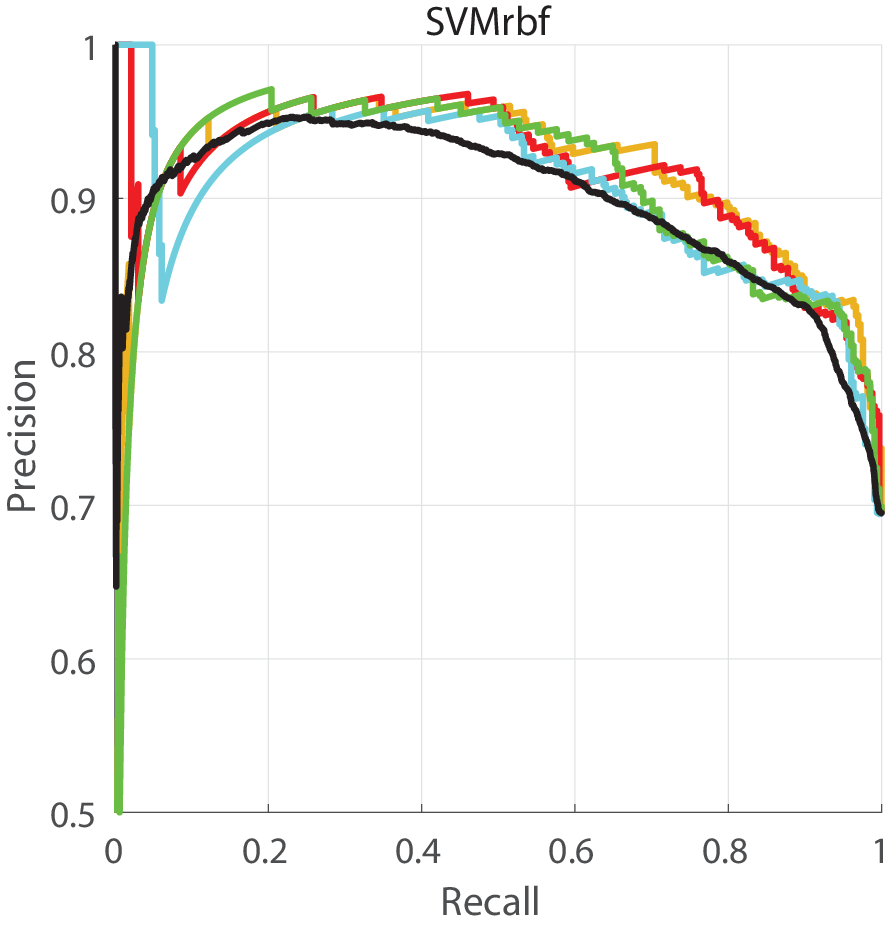}

\end{figure*} 

\begin{figure*}[h]
\caption{$F1$-score vs. threshold at different moments before the overtake maneuver starts.}
\label{fig:F1curves}
\centering

\includegraphics[width=0.24\textwidth]{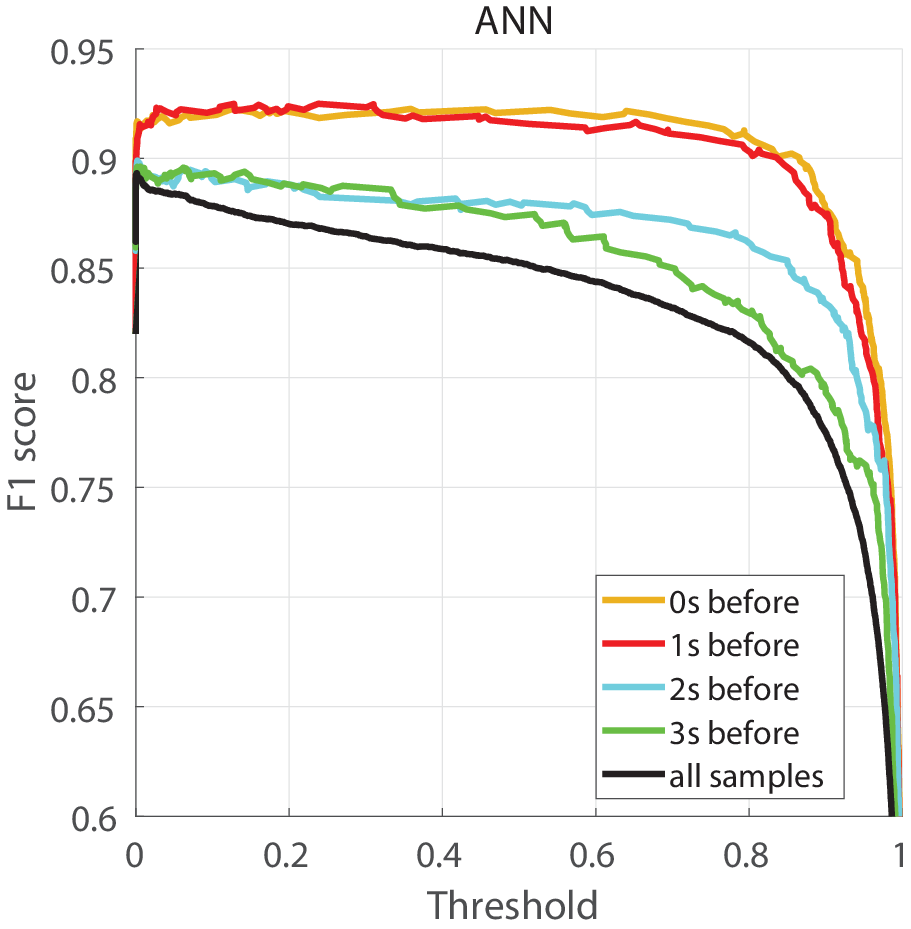}
\includegraphics[width=0.24\textwidth]{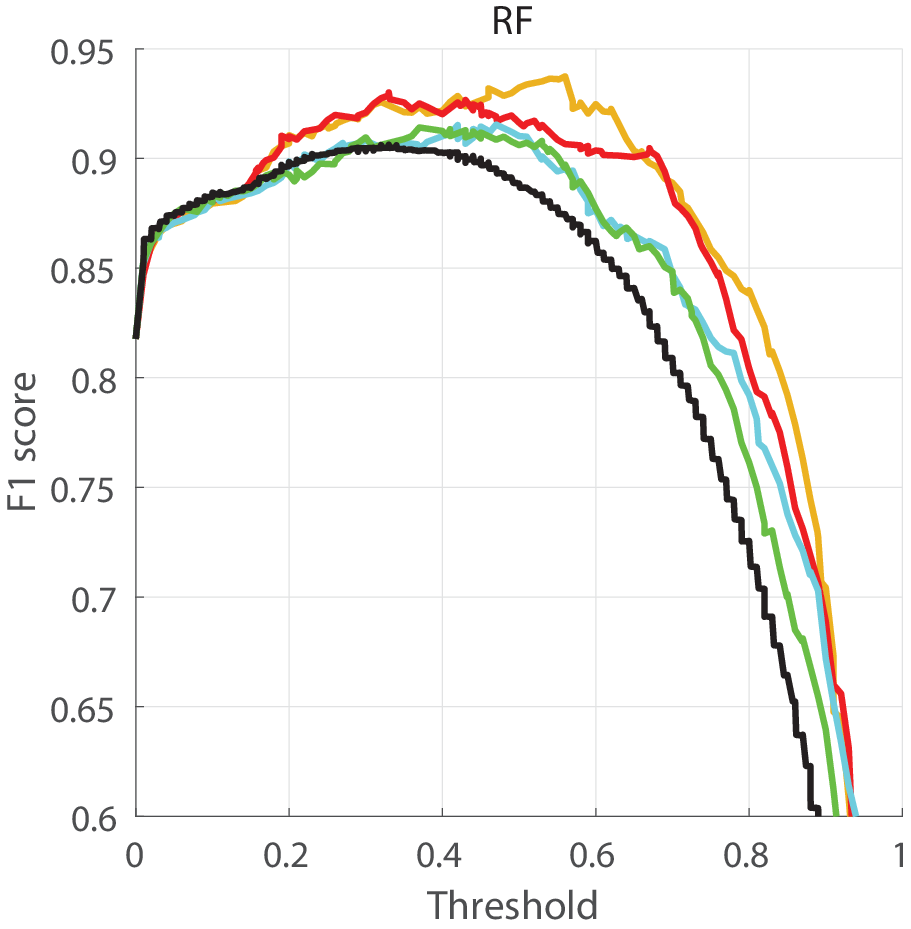}
\includegraphics[width=0.24\textwidth]{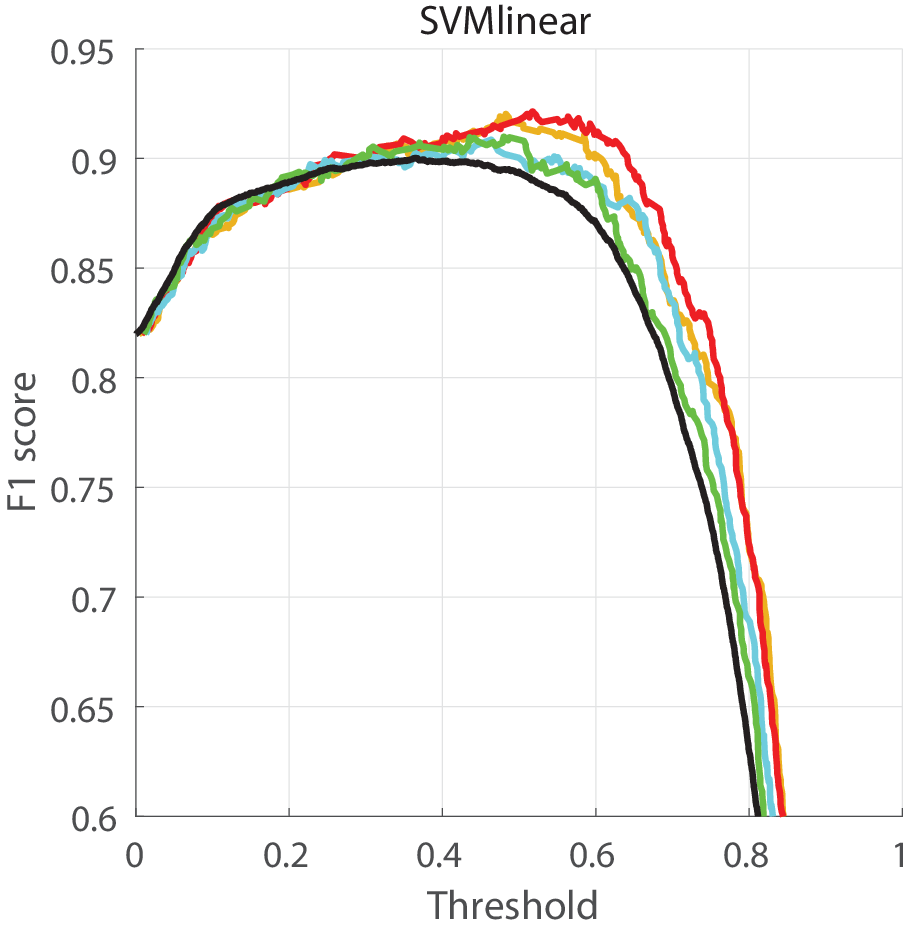}
\includegraphics[width=0.24\textwidth]{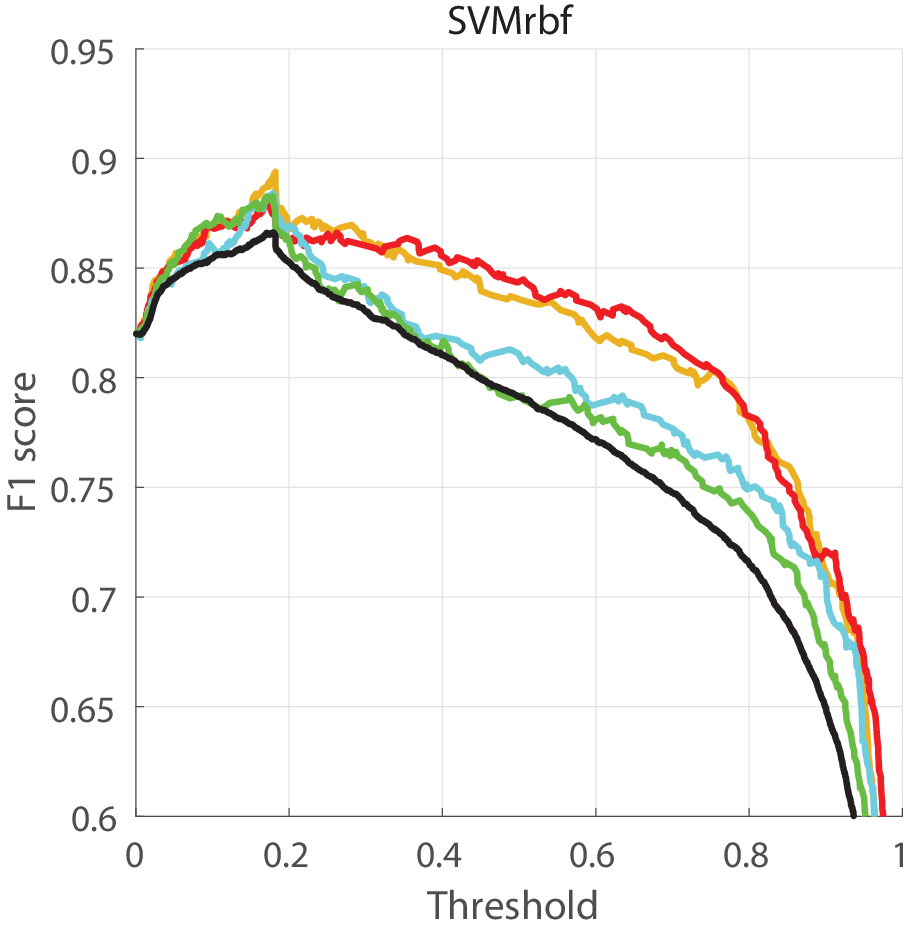}

\end{figure*}

\section{Experimental framework}
\label{sec:ef}

\subsection{Database}

Our database consists of data from 3 real operating trucks normally driving around Europe, provided by Volvo Group participating in this research.
The trucks are equipped with a data logger that captures CAN signals at 10 Hz.
The signals employed in this work include:

\begin{enumerate}
    \item Position of the accelerator pedal
    \item Distance to the vehicle ahead
    \item Speed of the vehicle ahead
    \item Relative speed difference between the vehicle and the left wheel 
    \item Vehicle speed
    \item Vehicle lateral acceleration
    \item Vehicle longitudinal acceleration
    \item Lane change status of the vehicle
    \item Status of the left turn indicator
    \item Status of the right turn indicator
    
\end{enumerate}

To avoid running out of storage, the data logger is programmed to record only when a precondition trigger to detect potential overtakes is met.
Such trigger is activated based on certain signals reaching specific thresholds: signal 8 (active), signal 5 (more than 50 km/h), signal 2 (less than 200 m), and signal 4 (more than 0.1 km/h).
When the trigger is activated, the logger saves the CAN signals from 20 seconds before the trigger up to 45 seconds thereafter. 
Data also includes video from a camera placed in the dashboard looking ahead the vehicle. 
Afterwards, a person manually labels the files by watching the videos and determines if it is an overtake or not. 
With this procedure, we obtained 264 files labelled as no overtakes and 448 files as overtakes.

\subsection{Classifiers}

To detect overtakes, we have used 3 classifiers: Artificial Neural Networks (ANN), Random Forest (RF), and Support Vector Machines (SVM, with linear and rbf kernels). 
They are based on different strategies and are a popular choice in the related literature \cite{c7}.
An ANN consists of several interconnected neurons that are arranged in layers (i.e., input, hidden, and output layers). Nodes in one layer are interconnected to all nodes in the neighbouring layers. Two design parameters of ANNs are the number of intermediate layers and the amount of neurons per layers.
An extension of the standard classification tree algorithm, the RF algorithm is an ensemble method where the results of many decision trees are combined. This helps to reduce overfitting and to improve generalization capabilities. The trees in the ensemble are grown by using bootstrap samples of the data. 
Finally, SVM searches for an optimal hyperplane in a high dimensional space that separates the data into two classes. SVM
uses different kernel functions to transform data that can be used to form the hyperplane, such as linear, gaussian or polynomial.

In this work, the available files are cropped from -10 seconds to +1 around the precondition trigger, following \cite{c6}. 
At 10 Hz, this gives 110 samples per file.
The CAN signals are then analyzed via a sliding window of 1 second with 50\% overlap, resulting in 21 samples per file.
For signals 1-7 (non-categorical), we compute the mean and standard deviation of the samples inside the window \cite{c8}, whereas for signals 8-10 (categorical) we extract the majority value among the window samples.
All samples from overtake files are labelled as class1 (positive class or overtake), whereas all samples for no-overtake files are labelled as class0 (negative class or no-overtake).
The training data is balanced per class. It means that we check how many files of each class are available per truck, then we take the 70\% of the minimum. All other files are used for testing.
This results in the amount of files indicated in Table~\ref{tab0}.

Experiments are conducted using Matlab r2023b. All classifiers are left with the default values (ANN: one hidden layer with 10 neurons; RF: 100 decision trees), except:

\begin{itemize}
    \item ANN and SVM use standardization (subtract the mean, and divide by std of training data)
    \item The ANN iteration limit is raised to 1e6 (from 1e3) to facilitate convergence
    \item Similarly, the SVMrbf iteration limit is raised to 1e8 (from 1e6)
\end{itemize}

\begin{table}[h]

\caption{AUC-PR of the classifiers at different moments before the overtake manoeuvre starts ($t$ corresponds to the precondition
trigger, $t$-1 to one second earlier, and so on). The PR curves are shown in Figure~\ref{fig:PRcurves}. The row \textit{variation} shows the difference between RF+SVML and the best AUC (Area under the curve) of the RF and SVML classifiers. The bold number in each column indicates the results of the best individual classifier. If the fusion RF+SVML improves the best individual classifier, such a cell is also marked in bold.}

\label{tab1}

\resizebox{0.48\textwidth}{!}{\begin{tabular}{|l|l|l|l|l|l|l|}
\hline
 &   &  &  &  & all \\
 \textbf{classifier} &  t & t-1 & t-2 & t-3 & samples\\
\hline
ANN & 0.931 & 0.914 & 0.907 & 0.890 & 0.880 \\ 
RF & 0.896 & 0.885 & 0.890 & 0.900 & 0.902 \\
SVML & \textbf{0.952} & \textbf{0.950} & \textbf{0.946} & \textbf{0.949} & \textbf{0.951} \\
SVMrbf & 0.914 & 0.915 & 0.903 & 0.906 & 0.897 \\ \hline
RF+SVML & \textbf{0.981} & \textbf{0.981} & \textbf{0.975} & \textbf{0.974} & \textbf{0.973} \\
\textit{variation} & +0.029 & +0.031 & +0.029 & +0.025 & +0.022 \\
\hline

\end{tabular}}
\end{table}

\begin{table*}[!hbt]
\caption{Precision, recall and F1-score (values in \%) of the classifiers at different moments before the overtake manoeuvre starts ($t$ corresponds to the precondition trigger, $t$-1 to one second earlier, and so on). We use the threshold (th) which gives the maximum F1-score (Figure~\ref{fig:F1curves}). The row \textit{variation} shows the difference between RF+SVML and the best of the RF and SVML classifiers. The bold number in each column indicates the results of the best individual classifier. If the fusion RF+SVML improves the best individual classifier, such a cell is also marked in bold.}\label{tab2}
\centering
\small
    \begin{adjustbox}{max width=\textwidth}
    \begin{tabular}{|l||l|l|l|l||l|l|l|l||l|l|l|l||l|l|l|l||l|l|l|l|}
    \hline
         &  \multicolumn{4}{|c||}{t}        
         & \multicolumn{4}{|c||}{t-1} 
         & \multicolumn{4}{|c||}{t-2} 
         & \multicolumn{4}{|c||}{t-3} 
         & \multicolumn{4}{|c|}{all samples}\\ \hline
         
        \textbf{classifier} & Prec & Rec & F1 & th & Prec & Rec & F1 & th & Prec & Rec & F1 & th & Prec & Rec & F1 & th & Prec & Rec & F1 & th \\ \hline
        
        ANN & 90.12 & 94.51 & 92.26 & 0.13 & 91.12 & 93.90 & 92.49 & 0.24 & 84.27 & \textbf{96.34} & 89.90 & 0.00 & 84.97 & \textbf{94.82} & 89.63 & 0.01 & 84.52 & 94.72 & 89.33 & 0.00 \\ \hline
        
        RF & \textbf{95.05} & 92.47 & \textbf{93.74} & 0.56 & 88.35 & \textbf{98.19} & \textbf{93.01} & 0.33 & \textbf{91.82} & 91.27 & \textbf{91.54} & 0.47 & 88.45 & 94.58 & \textbf{91.41} & 0.37 & 86.84 & \textbf{94.75} & \textbf{90.62} & 0.33 \\ \hline
        
        SVML & 90.80 & 93.29 & 92.03 & 0.48 & \textbf{91.32} & 92.99 & 92.15 & 0.52 & 89.38 & 92.38 & 90.85 & 0.46 & \textbf{91.13} & 90.85 & 90.99 & 0.48 & \textbf{87.11} & 93.13 & 90.02 & 0.36 \\ \hline
        
        SVMrbf & 83.38 & \textbf{96.34} & 89.39 & 0.18 & 81.94 & 95.43 & 88.17 & 0.17 & 83.29 & 94.21 & 88.41 & 0.18 & 83.06 & 94.21 & 88.29 & 0.18 & 82.25 & 91.49 & 86.63 & 0.18  \\ \hline \hline
        
        
        RF+SVML & \textbf{97.12} & 91.27 & \textbf{94.10} & 0.59 & \textbf{95.91} & 91.87 & \textbf{93.85} & 0.57 & \textbf{97.00} & 87.65 & \textbf{92.09} & 0.59 & \textbf{96.99} & 87.35 & \textbf{91.92} & 0.59 & \textbf{92.99} & 88.45 & \textbf{90.66} & 0.51  \\ \hline
        
        \textit{variation} & +2.07 & -2.03 & +0.36 & ~ & +4.59 & -6.33 & +0.84 & ~ & +5.18 & -4.73 & +0.55 & ~ & +5.86 & -7.23 & +0.51 & ~ & +5.88 & -6.30 & +0.04 & ~ \\ \hline
        
    \end{tabular}
    \end{adjustbox}
\end{table*}

\begin{table*}[!hbt]
\caption{$TPR/TNR$ of the classifiers at different moments before the overtake manoeuvre starts ($t$ corresponds to the precondition trigger, $t$-1 to one second earlier, and so on). 
The row \textit{variation} shows the difference between RF+SVML and the best of the RF and SVML classifiers. The bold number in each column indicates the results of the best individual classifier. If the fusion RF+SVML improves the best individual classifier, such a cell is also marked in bold.}

\label{tab3}

    \centering
    \small
    \begin{adjustbox}{max width=0.8\textwidth}
    \begin{tabular}{|l||l|l||l|l||l|l||l|l||l|l|}
    \hline

        \textbf{classifier} & \multicolumn{2}{|c||}{t} & \multicolumn{2}{|c||}{t-1} & \multicolumn{2}{|c||}{t-2} & \multicolumn{2}{|c||}{t-3} & \multicolumn{2}{|c|}{all samples} \\ \hline
        ~ & TPR & TNR & TPR & TNR & TPR & TNR & TPR & TNR & TPR & TNR \\ \hline
        
        ANN & 94.51\% & 76.39\% & 93.90\% & 79.17\% & \textbf{96.34\%} & 59.03\% & \textbf{94.82\%} & 61.81\% & 94.72\% & 60.48\%  \\ \hline
        
        RF & 92.47\% & \textbf{89.19\%} & \textbf{98.19\%} & 70.95\% & 91.27\% & \textbf{81.76\%} & 94.58\% & 72.30\% & \textbf{94.75\%} & 67.79\% \\ \hline
        
        SVML & 93.29\% & 78.47\% & 92.99\% & \textbf{79.86\%} & 92.38\% & 75.00\% & 90.85\% & \textbf{79.86\%} & 93.13\% & \textbf{68.62\%} \\ \hline
        
        SVMrbf & \textbf{96.34\%} & 56.25\% & 95.43\% & 52.08\% & 94.21\% & 56.94\% & 94.21\% & 56.25\% & 91.49\% & 55.03\% \\ \hline \hline
        
        RF+SVML & 91.27\% & 93.92\% & 91.87\% & 91.22\% & 87.65\% & 93.92\% & 87.35\% & 93.92\% & 88.45\% & 85.04\% \\ \hline
        
        \textit{variation} & -2.03\% & 4.73\% & -6.33\% & 11.36\% & -4.73\% & 12.16\% & -7.23\% & 14.06\% & -6.30\% & 16.42\% \\ \hline
    \end{tabular}
   \end{adjustbox} 
\end{table*}

\section{Results}
\label{sec:res}

In Figure~\ref{fig:boxplots}, we present the boxplots of the decision scores of each classifier towards the two classes. 
Notice that the classifiers are set to produce the probability that a sample belongs to a specific class (i.e. belonging to [0,1]).
It can be observed that the output probability of class1 (overtake) usually increases as the precondition trigger approaches ($x$-axis=0), whereas class0 keeps a stable or oscillating probability, depending on the classifier.
Thus, from the right plot of Figure~\ref{fig:boxplots}, it can be seen that it will be easier to detect overtakes closer to the trigger.

We then report in Figure~\ref{fig:PRcurves} the Precision-Recall (PR) curves of the classifiers at different moments before the precondition trigger. 
We also provide results considering all samples of the files at any given instant from -10 seconds to +1 seconds around the trigger.
Table~\ref{tab1} gives the AUC (Area under the curve) values.
Precision measures the proportion of detected positives which are actually overtakes, quantified as:

\begin{equation}
P = \frac{{TP}}{{TP + FP}}   
\label{eq:P}
\end{equation} 

Recall, on the other hand, measures the amount of overtakes that are actually detected, as:

\begin{equation}
R = \frac{{TP}}{{TP + FN}} 
\end{equation}

A summarizing measure of P and R is the $F1$-score, defined as:

\begin{equation}
    F1 = 2\frac{{P \times R}}{{P + R}}
\end{equation}

Figure~\ref{fig:F1curves} provides the $F1$-score for different values of the threshold applied to the decision scores.
The mentioned curves confirm the observation that “the closer to the trigger, the better”.
It can be seen that orange curves (0s before the trigger) and red curves (1s before the trigger) usually appear above the others.
The black curves (which use samples in the entire range of -10 seconds to +1 seconds around the trigger) always show the worst behaviour. This confirms that samples earlier than 3 seconds before the trigger actually provide worse detection capabilities, making more difficult to predict overtakes earlier.

We then select the threshold of each classifier and moment that provides the highest $F1$-score.
Table~\ref{tab2} reports $P$, $R$ and $F1$, whereas Table~\ref{tab3} reports the true positive rate ($TPR$) and false positive rate ($FPR$), calculated as follows:

\begin{equation}
TPR = \frac{{TP}}{{TP + FN}}
\end{equation}

\begin{equation}
TNR = \frac{{TN}}{{TN + FP}}
\end{equation}

\begin{figure*}[h]
\caption{Graphical plot of $TPR/TNR$ at different moments before the overtake manoeuvre starts ($t$ corresponds to the precondition trigger, $t$-1 to one second earlier, and so on).}
\label{fig:TPR-TNR}
\centering

\includegraphics[width=0.48\textwidth]{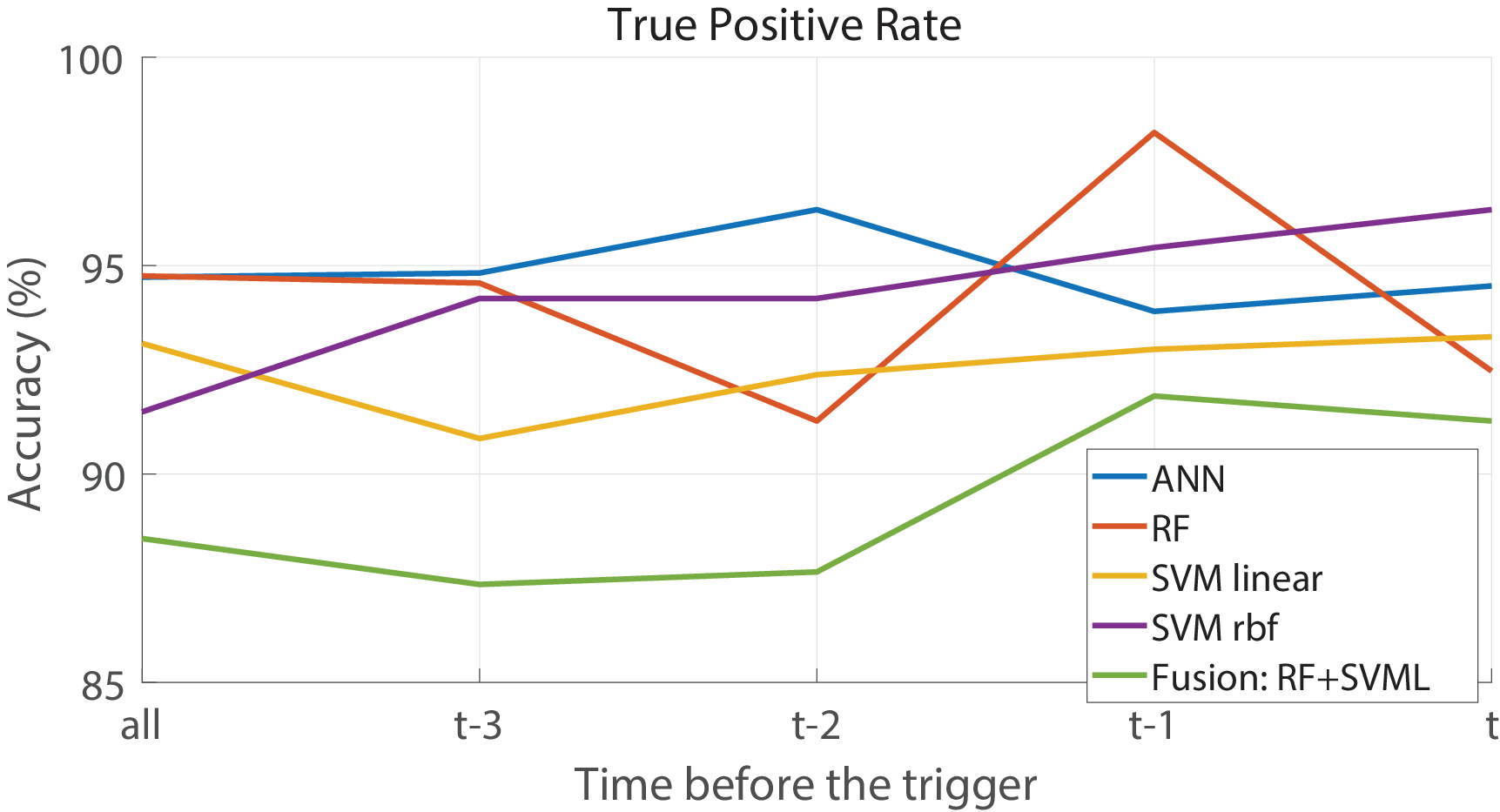}
\includegraphics[width=0.48\textwidth]{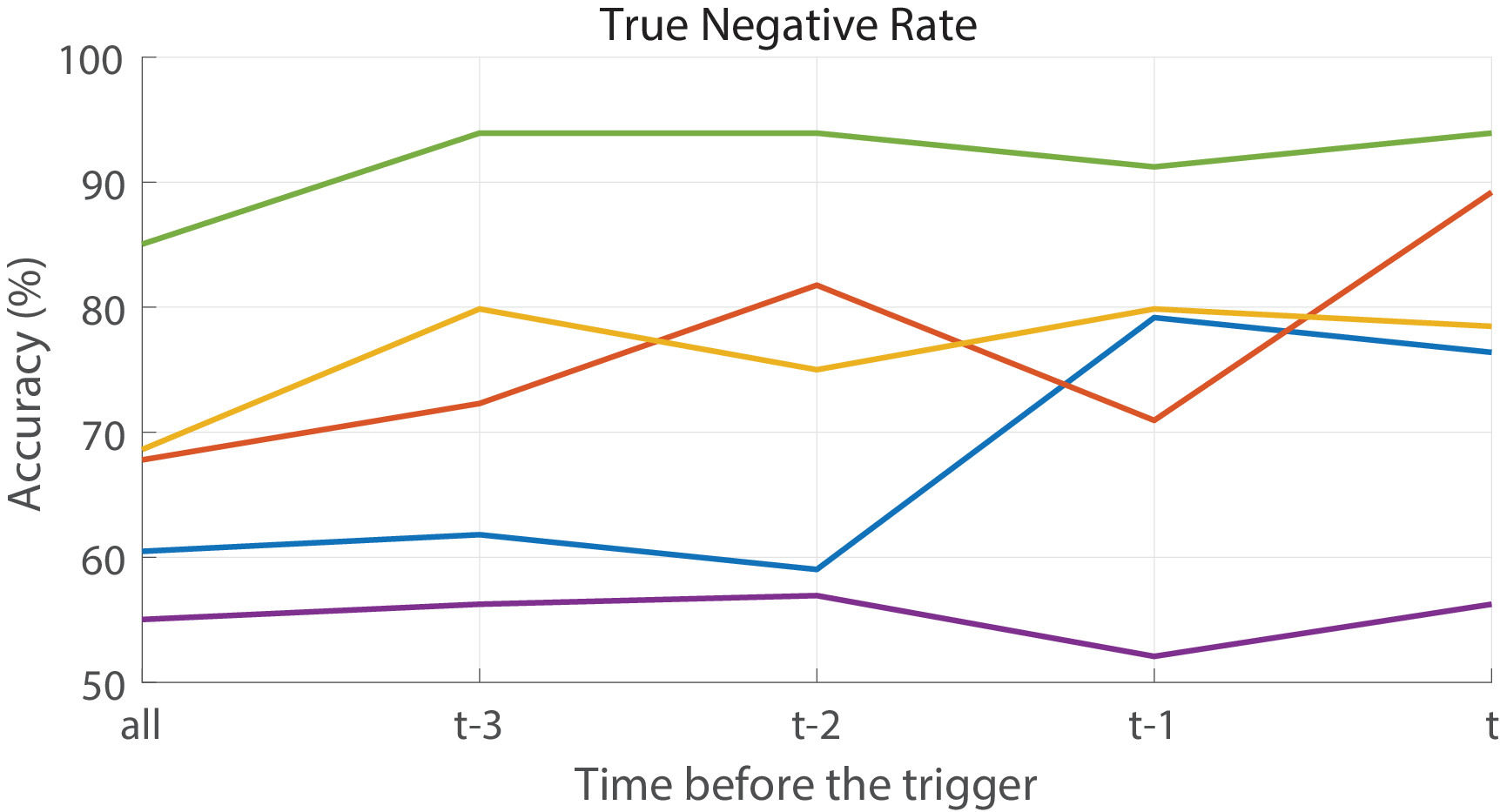}

\end{figure*}

$TPR$ measures the amount of overtakes that are actually labelled as overtakes, whereas $TNR$ measures the amount of no overtakes that are actually labelled as no overtakes.
Notice that $TPR=R$.
The bold values in the tables show that Random Forest (RF) usually stands out as the best individual classifier, consistently obtaining the highest $F1$ at any given moment in time.
To better observe the evolution of $TPR/TNR$, we graphically show in Figure~\ref{fig:TPR-TNR} their values at different moments before the trigger.
$TPR$ stands above 90\% for all classifiers, even when using all samples within 10 seconds before the trigger, meaning that actual overtakes can be well detected. Random Forest gives the best accuracy ($>$98\% at t-1), although its performance is somehow more erratic across time.
ANN is the classifier with the most stable $TPR$ at any time (above 94\%). 
Interestingly, not all classifiers have their best $TPR$ at $t$ (exact moment of the trigger). 
As it was observed in the boxplots of Figure~\ref{fig:boxplots}, the score towards the positive class (right columns) tends to decrease abruptly exactly at the trigger. This could be because the window is capturing a portion of samples after the trigger, which is shown to actually be detrimental to the detection.
Regarding $TNR$ (left plot of Figure~\ref{fig:TPR-TNR}), its values can diminish to as low as the 50-60\% range, meaning that a substantial percentage of no overtakes would be actually labelled as overtakes.
Here, RF and ANN show better numbers ($TNR$ above 70-80\%).
Also, in this case, it is actually observed that the farther away from the trigger, the lower the $TNR$.

From the results above, we observe that $TNR$ is not as high, so the classifiers are not as good in classifying no overtakes.
Also, ANN and SVMrbf show some strange behaviour, such as that the threshold of maximum $F1$ is too low (Table \ref{tab2}), or the P-R curves are too “shaky”.
This suggests that the default values of these classifiers may not be the best choice.
We thus take RF and SVM linear further and fuse their output scores by taking their mean.
The AUC, $P$, $R$, $F1$, $TNR$ and $TPR$ of the fusion have been also provided in Tables~\ref{tab1}-\ref{tab3}.
It can be observed that AUC, Precision, $F1$ and True Negative Rates improve for all moments before the trigger.
On the other hand, Recall and True Positive Rates are seen to decrease.
The observed effect of the fusion is that the ability to classify no overtakes is increased, at the cost of reducing overtake detection capabilities.
However, the increase in $TNR$ is much bigger than the decrease in $TPR$ (Tables~\ref{tab3}).
Overall, the fusion provides a more balanced accuracy of these two metrics, situating them beyond 91\%. 
For example, at t-1 or earlier, $TNR$ was below 80\%, but after the fusion, as early as 3 seconds before the trigger, both classes have an accuracy of 87\% or higher. 
Such stability and well-balanced accuracy can also be observed in Figure~\ref{fig:TPR-TNR}.

\section{Conclusions}
\label{sec:con}

We demonstrate the suitability of CAN bus data to detect overtakes in trucks.
We do so via traditional widely used classifiers \cite{c7}, including Artificial Neural Networks (ANN), Random Forest (RF), and Support Vector Machines (SVM).
To the best of our knowledge, we are the first to apply machine learning techniques for overtake detection of trucks from CAN bus data.
The classifiers employed performed well for the overtake class (TPR $\ge$ 93\%), although their performance is not as good in the no overtake class.
With the help of classifier fusion, the accuracy of the later class is observed to increase, at the cost of some decrease in the overtake class. Overall, the fusion balances TPR and TNR, providing more consistent performance than individual classifiers.

As future work, we are exploring the optimization of classifiers beyond their default values \cite{c9}. 
Parameters like the size of the sliding window employed or the time ahead of the precondition
trigger are also subject to discussion in the literature \cite{c1,c7}.
There is the possibility of capturing large amounts of continuous unlabeled data from Volvo Group participating in this research.
We are also considering the improvement of the developed classifiers by training them on a larger dataset obtained via pseudo-labeled data \cite{c10}, for example, selecting samples with high prediction probability as given by the classifiers trained with labelled data.
This would avoid the time-consuming manual labelling issue.
A bigger dataset would also enable the use of data-hungry popular models such as Long Short-Term Memory (LSTM) networks \cite{c11}.

\section*{Acknowledgements}

The authors thank the BigFun project of the Swedish Innovation Agency (VINNOVA) for funding their research.

\end{document}